\documentclass{article}

\usepackage{scml2025}
\usepackage{multicol}
\usepackage[utf8]{inputenc} 
\usepackage[T1]{fontenc}    
\usepackage{hyperref}       
\usepackage{url}            
\usepackage{booktabs}       
\usepackage{amsfonts}       
\usepackage{nicefrac}       
\usepackage{microtype}      
\usepackage{enumitem}
\usepackage{lipsum}         
\usepackage{graphicx}
\usepackage{doi}
\usepackage{wrapfig}
\usepackage{float}
\usepackage{amsmath}
\usepackage{cleveref}       

\title{Evidential Physics-Informed Neural Networks}
\renewcommand{\shorttitle}{E-PINN}

\date{}

\usepackage{authblk}

\author[1]{
	Hai Siong Tan}
\author[2]{
Kuancheng Wang}
\author[3]{Rafe McBeth}
\affil[1]{Gryphon Center for Artificial Intelligence and Theoretical Sciences, Singapore}
\affil[2]{Georgia Institute of Technology, Atlanta, GA, USA}
\affil[3]{University of Pennsylvania, Perelman School of Medicine, Department of Radiation Oncology, PA, USA}

\begin{document}
\maketitle

\begin{abstract}
We present a novel class of Physics-Informed Neural Networks that is formulated based on the principles of Evidential Deep Learning, where the model incorporates uncertainty quantification by learning parameters
of a higher-order distribution. 
The dependent and trainable variables of the PDE residual loss and data-fitting loss terms are recast as functions of the hyperparameters of an evidential prior distribution. Our model is equipped with an information-theoretic regularizer that contains the Kullback-Leibler divergence
between two inverse-gamma distributions characterizing predictive uncertainty. 
Relative to Bayesian-Physics-Informed-Neural-Networks, our framework appeared to exhibit higher sensitivity to data noise, preserve boundary conditions more faithfully and yield empirical coverage probabilities closer to nominal ones. Toward examining its relevance for data mining in scientific discoveries, we demonstrate how to apply our model to inverse problems involving 1D and 2D nonlinear differential equations. 
\end{abstract}

\begin{multicols}{2}
\section{Introduction}
Physics-Informed Neural Networks (PINNs)
furnish a class of scientific machine learning frameworks that can harness observational data and the universal approximation capability of neural networks \cite{universal} to solve partial differential equations (PDE) and their related inverse problems.
It is essentially characterized by a linear combination of loss functions formulated such that they drive the model towards fitting the empirical datasets while also adhering to the governing PDEs (equipped with some set of boundary/initial conditions). Over the years, PINNs have shown excellent results in terms of solving not only forward problems \cite{Lagaris_1998, pinn, raissi2017},
but also inverse ones \cite{cuomo2022, hao2023, Ameya, kim, zhou2024} where we use PINNs to determine unknown scientific parameters in the PDEs that are assumed to model the observational data, enabling deep learning techniques to be used as a powerful tool for scientific discovery. 

In this paper, we present a novel form of PINNs designed such that uncertainty quantification \cite{abdar} is naturally embedded within the model architecture and its loss function. In our formulation, uncertainty estimates of both the dependent variables of the PDE and its unknown parameters are obtained immediately upon completion of model training which is implemented in a typical gradient-descent based approach. This is in contrast to other forms of methods such as Bayesian-based techniques which rely on Monte-Carlo sampling to estimate output variance. Our model is largely inspired by the principles of \emph{Evidential Deep Learning} (EDL) introduced in the seminal works \cite{amini, sensoy} where an \emph{evidential}, higher-order set of priors are placed over likelihood functions for the outputs, and model training is then used to infer hyperparameters of this distribution. Uncertainty estimates can be deduced straightforwardly as they are closed-form functions of the learned hyperparameters. To our knowledge, although EDL has been implemented in various types of model architectures (e.g. U-Nets \cite{Jones, Zou}, FNN (\cite{amini}), LeNET \cite{sensoy}) it hasn't been synthesized with PINNs. In our work here, we propose how a hybrid of PINN and EDL principles can be performed, and present a novel information-theoretic regularizer for EDL that is also applicable for more general regression tasks. 

As reviewed in \cite{cuomo2022}, one of the most recent state-of-the-art proposals for uncertainty quantification in PINNs is the highly interesting Bayesian-PINN (B-PINN) of Yang et al. in \cite{bpinn}. Thus, for comparison, we validated our model against the inverse PDE problems studied in \cite{bpinn}. Specifically, we applied our model to 1D and 2D nonlinear reaction-diffusion-type systems with the training datasets being doped with heterogeneous noise. We found that relative to B-PINN, our model inherited uncertainty estimates which were significantly more sensitive 
to the added data noise, and correlated well with prediction errors for out-of-distribution data. It adhered to the boundary conditions more faithfully in the examples we examined relative to other frameworks. We call our class of models \emph{Evidential PINN} (or E-PINN for short), and hope that it adds to the uncertainty quantification toolkit for scientific machine learning at large.

\section{Theoretical basis of our model}
\label{Prelim}
A distinctive feature of PINNs is that the PDE residual is included as an essential component of the loss function, guiding the model training towards adhering to the PDE in an unsupervised manner. Consider the case where the PDE contains an unknown parameter $\kappa$. We lift it to be a random variable equipped with a density function $\mathcal{P}(\kappa; \vec{\mu}_\kappa)$ where $\vec{\mu}_\kappa$ collectively denotes its moments. In the most general case, $\kappa$ is not separable from the PDE residual which we denote as 
\begin{equation}
\label{residual}
\mathcal{L}\left( 
u(\vec{x}), \partial u, \vec{x}, \kappa
\right) = 0,
\end{equation}
where $\partial u$ denotes all differential operators acting on the dependent variable $u$. Typical PINN formalism asserts $\mathcal{L}^2$ to be part of the overall loss function. In our model, we replace it by a marginalized version, integrating the square of \eqref{residual} over the domain of $\kappa$ ($\mathcal{D}_\kappa$) to obtain
\begin{equation}
\label{residual_gen}
\mathcal{G}\left( u, \partial u, \vec{x}, \vec{\mu}_\kappa \right) \equiv
\int_{\mathcal{D}_\kappa} d\kappa
\,\,\,
\mathcal{L}^2 \cdot \mathcal{P}(\kappa; \vec{\mu}_\kappa ) = 0, 
\end{equation}
with $\mathcal{G}$ being the weighted PDE residual, and $\vec{\mu}_\kappa$ are now learnable parameters instead of $\kappa$. The usual PINN formalism proceeds from here, yet an explicit form for $\mathcal{P}(\kappa; \vec{\mu}_\kappa)$ has to be specified, and $\mathcal{G}$ is generally dependent on the chosen $\kappa$ prior. For concreteness, in this paper, we consider the case where $\kappa$ is separable from other terms within $\mathcal{L}$ in the following sense.  
\begin{equation}
\label{simple_L}
\mathcal{L}  = \mathcal{L}_1\left(
u, \partial u, \vec{x} \right) + 
\kappa \, \mathcal{L}_2\left(
u, \partial u, \vec{x} \right).
\end{equation}
Marginalizing over $\kappa$ via \eqref{residual_gen}, we then obtain 
\begin{equation}
\label{simple_resi}
\mathcal{G}\left( u, \partial u, \vec{x}, \vec{\mu}_\kappa \right)  = \mathcal{L}_1^2
+ \mathcal{L}_2^2 \left( \text{Var}(\kappa) + \overline{\kappa}^2 \right)
+ \overline{\kappa}\left( 2\mathcal{L}_1 \mathcal{L}_2  \right),
\end{equation}
where $\vec{\mu}_\kappa$ = $\{\overline{\kappa}, \text{Var}(\kappa)\}$ is part of the list of learnable parameters. Note that 
\eqref{simple_resi} is agnostic to the exact form of the prior $\mathcal{P}(\kappa;\vec{\mu}_\kappa )$, and that it reduces to the usual PDE residual loss of PINNs in the limit $\text{Var}(\kappa) \rightarrow 0$
with $\overline{\kappa} \rightarrow \kappa$. Despite its simplicity, the PDE residual \eqref{simple_resi} for $\mathcal{L}$ is applicable for many inverse PINN problems studied in previous literature (e.g. \cite{baty2024, Chen_2020, bpinn}). 
It can be encapsulated within traditional Monte-Carlo Dropout \cite{gal} and Deep Ensemble \cite{Egele, DE, lak} methods to yield uncertainty estimates for $u(\vec{x})$ apart from $\kappa$. 

We now incorporate the fundamental principles of EDL \cite{amini, sensoy}
in PINN and the residual loss \eqref{simple_resi}. We first assume that the observed datapoints of $u$ have been drawn i.i.d. from Gaussian profiles. Placing Gaussian and inverse-gamma priors on the means and variances respectively, the posterior distribution adopts the form of a 4-parameter normal-inverse-gamma distribution. 
\begin{equation}
\label{NIG}
\tilde{p}\left(\overline{u}, \sigma^2_u 
\vert \gamma, \nu, \alpha, \beta \right) = 
\frac{\beta^\alpha \sqrt{\nu}}
{\Gamma (\alpha) \sqrt{2\pi \sigma^2_u}}
e^{- \frac{2\beta + \nu(\gamma - \overline{u})^2}{2\sigma^2_u}}.
\end{equation}
This assertion leads to simple closed-form formulas for the prediction means and uncertainties as follows. 
\begin{equation}
\label{output}
\mathbb{E}[\overline{u}] = \gamma, \,\,\,
\sigma^2_p = \frac{\beta}{\alpha - 1}
\left( 
1 + \frac{1}{\nu}
\right),
\end{equation}
where we note that the expectation value $\mathbb{E}$ is defined w.r.t. $\tilde{p}$ in \eqref{NIG}, and that the first and second components within the bracket are the aleatoric and epistemic uncertainties defined as 
$\mathbb{E}[\sigma^2_u], \text{Var}[\overline{u}]$ respectively. 
In EDL, the learnable parameters are the hyperparameters $\{\gamma, \alpha, \beta, \nu \}$. Fitting data to the evidential model requires not the usual mean-squared-error loss term but rather the negative logarithm of the marginal likelihood obtained after integrating the first-order Gaussian profiles over the measure described by $\tilde{p}$ in \eqref{NIG}. 
\begin{equation}
\mathcal{L}_{edl}
= 
\log \left( (u_i - \gamma)^2\nu + \Omega \right)^{( \alpha + \frac{1}{2})} + 
\log \frac{
\sqrt{\pi/\nu}\, \Gamma(\alpha)
}{
\Omega^\alpha \Gamma(\alpha + \frac{1}{2} )
},  
\end{equation}
where $\Omega \equiv 2\beta(1+\nu)$ and $u_i$ refer to the observed datapoints for the dependent variable $u$. In \cite{amini}, the authors proposed a regularizer designed to minimize evidence on incorrect predictions that reads 
$\mathcal{L}_{R} = |u_i - \gamma|(2\nu + \alpha)$. This was shown in \cite{amini, Tan_2} to work well, yet it is independent of $\beta$ which controls the overall uncertainty scale. Here we propose a refinement of this original EDL regularizer by multiplying it to the Kullback-Leibler divergence \cite{thomas, Kullback} between the distribution described by \eqref{NIG} and a weakly-informative prior, its final form being
(see Appendix A for a detailed derivation)  \begin{equation}
\label{kl}
\mathcal{L}_{kl} = |u_i - \gamma|(2\nu + \alpha) \left(
\log \frac{\beta^\alpha_r \, \Gamma(\alpha)} {\beta^\alpha} + \gamma_e (\alpha - 1)
+ \frac{\beta- \beta_r}{\beta_r}
\right),
\end{equation}
where $\gamma_e$ is the Euler-Mascheroni constant. In \eqref{kl}, 
$\beta_r$ is a reference $\beta$ value which is assumed to describe a weakly informative prior for larger deviations from the observed dataset. We set it to be of the same order-of-magnitude as the maximal target variable over the PDE domain.
Empirically, we found this regularizer to perform better than $\mathcal{L}_R$ that was used in \cite{amini, Tan_2}. Theoretically, it aligns with the goal of guiding distributions for erroneous predictions to inherit weak priors with higher uncertainties. 

To include a suitable PDE residual in the EDL framework, we take $\mathcal{G}$ as defined in \eqref{simple_resi}, but with $u \rightarrow \gamma$. Physically, this can be interpreted as the dependent variable being intrinsically stochastic in nature and the PDE should be understood as modeling the dynamics of its mean $\gamma$. The base model output variables are now represented by $\{\gamma, \alpha, \beta, \nu \}$ (instead of $u$) and thus the Dirichlet/Neumann boundary conditions should be expressed in terms of the mean $\gamma$. For the case studies here, we adopted the soft constraint approach for boundary conditions which were thus included as part of the training dataset. Our overall loss function for E-PINN reads
\begin{equation}
\label{loss}
\mathcal{L}_{epinn} =
\lambda_{d} \mathcal{L}_{edl}
+ \lambda_{kl} \mathcal{L}_{kl}
+ \lambda_{r} \mathcal{G}\left( \gamma, \partial \gamma, \vec{x}, \overline{\kappa}, \text{Var}(\kappa) \right),
\end{equation}
where $\lambda_d, \lambda_{kl}, \lambda_r $ are the weight coefficients for the data (EDL) loss term, the regularizer term and PDE residual term respectively. In \eqref{loss}, we restore the arguments of $\mathcal{G}$ to emphasize that the unknown parameter $\kappa$ is now represented by the trainable parameters $\overline{\kappa}, \text{Var}(\kappa)$. Their learning is coupled to the output variables $\{ \gamma, \alpha, \beta, \nu \}$ which yield the model's prediction and its associated uncertainty via \eqref{output}.

\section{Methodology}
We validated our model against the same 1D and 2D inverse problems presented in the seminal paper \cite{bpinn} where B-PINNs were formulated. They are nonlinear differential equations with source terms designed such that the exact solutions are simple products of sinusoidal functions, enabling a more convenient analysis of the neural networks' effectiveness. In contrast
to \cite{bpinn} however, we doped the training dataset with a more heterogeneous noise as we wish to scrutinize the degree to which the uncertainty distributions align with those of the inserted noise. The differential equations are of the same class as diffusion-reaction equations with non-linear source terms. 
\renewcommand{\labelenumi}{[\arabic{enumi}]} 

\begin{enumerate}
\item 1D nonlinear Poisson equation: 

$\lambda \frac{d^2u}{dx^2} + \kappa \tanh(u) = f_{1D}(x),\,\, u_e = \sin^3 (6x).$
\item 2D nonlinear diffusion-reaction PDE: 

$\lambda \nabla^2 u + \kappa \,u^2 = f_{2D}(x), \,\,
u_e = \sin(\pi x) \sin(\pi y).$
\end{enumerate}
Like in \cite{bpinn}, we set
$\lambda = 0.01$ and $\kappa$ to be the unknown PDE parameter to be inferred. In [1] and [2], the exact solutions $u_e$ and their corresponding source terms $f_{1D, 2D} (x)$ are such that we have $\kappa = \{0.7, 1 \}$ for the 1D and 2D equations respectively. For these problems, taking fully-connected multilayer perceptron (with a few hidden layers) to be the base models  turned out to be sufficient for our case studies. The 1D and 2D problems required only three and two hidden layers respectively. The total number of iterations was $10^5$ epochs, and Adam optimizer \cite{Adam} was used with a learning rate of $10^{-3}$. All E-PINN models were implemented with PyTorch \cite{pinnPy} using its \emph{Autograd} engine \cite{autograd}. For the 1D problem, the optimal loss function parameters were found to be $\lambda_d = 1, \lambda_{kl} = 0.01, \lambda_r = 0.1$, whereas for the 2D problem, we used
$\lambda_d = 1, \lambda_{kl} = 0.005, \lambda_r = 0.05$.
As for Bayesian-PINNs, the algorithm of \cite{bpinn} was implemented in PyTorch using the Hamiltonian-Monte-Carlo method \cite{cobb, bpinn} to sample the posterior distributions. The number of samples used was of the order of $10^4$ with a burn-in ratio of about $0.2$.

\begin{figure}[H]
	\begin{center}		\includegraphics[width=0.48\textwidth]{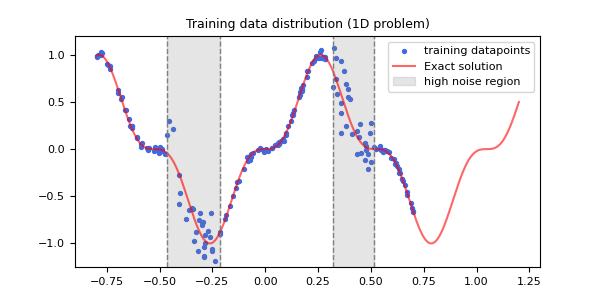}
	\end{center}
	\caption{\small Training data distribution for the 1D problem. The noise amplitude $\mathcal{A}(x)$ was set to be 0.2 for the shaded `high-noise' regions and 0.02 otherwise.}
	\label{fig:fig1}
\end{figure}

For the 1D problem, we took the number of observational and collocation data points to be 200 (random) and 500 (regularly spaced) respectively. The training dataset domain was $x \in (-0.8, 0.7)$, yet domain for testing data was extended to $x = 1.2$ for asessing various models on out-of-distribution data. 
The data was doped with a standard Gaussian noise with a spatially varying amplitude, i.e. $\mathcal{A}(x) \mathcal{N}(0,1)$ as depicted in Fig \ref{fig:fig1}. 

For the 2D problem, we also took 
the number of observational and collocation data points to be 200 and 500 respectively, with the domain being a square centered at the origin with length 2. In addition, we imposed a soft constraint realizing the Dirichlet condition along each of the boundary segments $\{ x=\pm 1, y=\pm 1 \}$ by including 50 points for each side of the square. We doped the training data with a standard Gaussian noise with a spatially varying amplitude $\mathcal{A}(x,y)$ that was maximal at the origin and vanishing along the boundary segments.
\begin{equation}
\label{noise2D_amp}
\mathcal{A}(x,y)
= 0.1 \cos \left( \frac{\pi x}{2} \right)
\cos \left( \frac{\pi y}{2} \right)
e^{-(x^2 + y^2)}.
\end{equation}
\begin{figure}[H]
	\begin{center}		\includegraphics[width=0.49\textwidth]{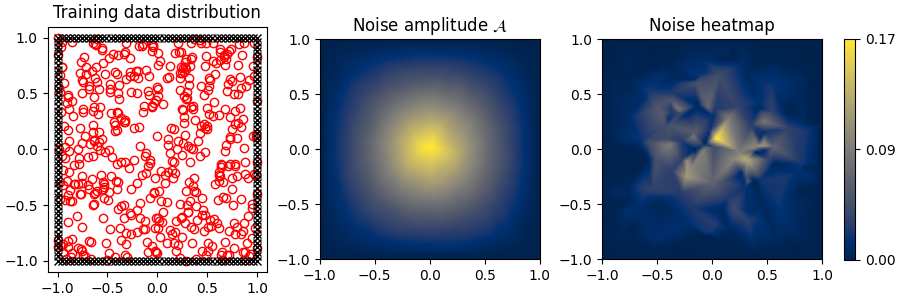}
	\end{center}
	\caption{\small Data and noise distributions for the 2D problem. The rightmost diagram show the noise generated by the stochastic function $\mathcal{A}(x,y) \mathcal{N}(0,1)$.}
	\label{fig:fig2}
\end{figure}
As a metric for assessing uncertainty quantification, we monitored the \emph{empirical coverage probability} (ECP) \cite{ECP2, ECP} defined as the proportion of samples of which deviations from the exact solution fell within the confidence band (we took it to be $\pm 1.96 \sigma_p$ for a $0.95$ CI). We also kept track of the Spearman's coefficients between $\sigma_p$ and (i)prediction errors ($\rho_e$) (ii)noise-induced errors in the training data ($\rho_n$). We compared our E-PINN model against B-PINN, Deep Ensemble (DE) method (defined by a set of 20 PINNs with different initial weights) and E-PINN using the original EDL regularizer of \cite{amini} rather than \eqref{kl} (denoted as EPINN(V) henceforth). 

\section{Results}
\begin{figure}[H]
	\begin{center}		\includegraphics[width=0.49\textwidth]{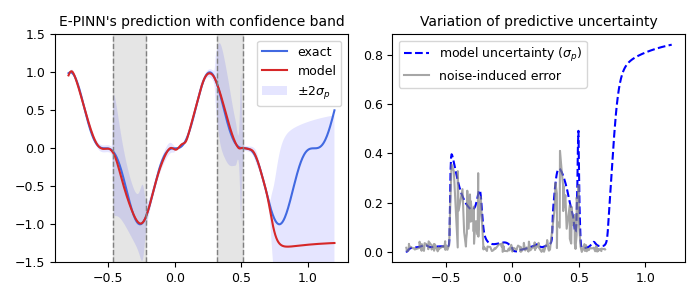}
	\end{center}
	\caption{\small E-PINN's predictions and uncertainty distributions (1D problem). }
	\label{fig:fig3}
\end{figure}

For the 1D problem, as visible in Fig. \ref{fig:fig3}, the predictive uncertainty $\sigma_p$ was larger in the high-noise regions and beyond the training domain, with the 0.95 confidence band encompassing both exact solution and model-predicted curve. Taking $\sigma_\kappa \equiv \sqrt{\text{Var}(\kappa)}$ to be the standard error, E-PINN predicted $\kappa \approx 0.71 \pm 0.03$, close to the actual value of $0.7$. The ECP value was 0.96 which was closest to the nominal value of 0.95 among all models. 
Compared to E-PINN(V), our model inferred a more accurate $\kappa$ and was better calibrated as measured by its ECP. Relative to B-PINN, it inherited stronger correlations with prediction errors $(\rho_e)$, noise-induced deviations of the training data from exact solution $(\rho_n)$ and had a better ECP score. The Deep Ensemble inferred $\kappa$ with a much smaller uncertainty than other models, but its ECP appeared to be problematic indicating a poor degree of calibration. Like B-PINN, the correlation of its uncertainty distribution with the noise of the data was weak ($\rho_n \sim 0.1$).
\begin{table}[H]
	\caption{\small Results for 1D ODE where $\kappa_{true} = 0.7$}
	\centering
    \small
	\begin{tabular}{ccccc}
		\toprule
$\,$    & \textbf{E-PINN} & E-PINN(V) & B-PINN & DE \\
		\midrule
$\kappa$ & \textbf{0.71}  & 0.74
& 0.69 & 0.6976 \\
$\sigma_\kappa$ & \textbf{0.03} & 0.03
& 0.03 & $10^{-4}$ \\
$\rho_e$ & \textbf{0.7} & 0.6 & 0.5 &
0.6 \\
$\rho_n$ & \textbf{0.7} & 0.7 & 0.2 &
0.1 \\
ECP & \textbf{0.96} & 0.79 & 0.87 &
0.38 \\
		\bottomrule
	\end{tabular}
	\label{tab:table}
\end{table}
For the 2D problem, we found E-PINN to exhibit similar advantages relative to other frameworks. It inferred $\kappa$ to be $1.00 \pm 0.01$, with a ECP value of $0.89$, and Spearman's coefficients $\rho_e \sim 0.7, \rho_n \sim 0.8$. The mean model error was $\sim 0.02$ while it was $\sim 0.002$ when restricted to boundary, showing strong adherence to the Dirichlet boundary condition. 
\begin{figure}[H]
	\begin{center}		\includegraphics[width=0.49\textwidth]{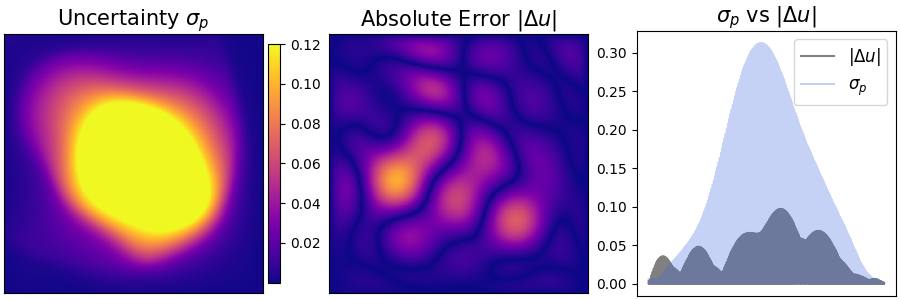}
	\end{center}
	\caption{\small E-PINN's predictions and uncertainty distribution heatmaps (2D problem). }
	\label{fig:fig4}
\end{figure}
For B-PINN, the mean model error was $ \sim 0.006$ while it was $\sim 0.01$ when restricted to boundary, showing weaker adherence to boundary conditions. E-PINN(V)'s prediction for $\kappa \approx 1.02 \pm 0.01$ did not quite cover the actual value of 1.00, and was weaker in its ECP value compared to E-PINN. The Deep Ensemble inferred $\kappa$ with an uncertainty much larger than other models, and its ECP value deviated most significantly from the nominal value. Uncertainty heatmaps and error plots for these models are relegated to Appendix \ref{sec:AppC}.
\begin{table}[H]
	\caption{\small Results for the 2D PDE where $\kappa_{true} = 1$. \emph{Abbrev:}  $\Delta u _b$ = mean model error evaluated along domain's boundary}
	\centering
    \small
	\begin{tabular}{ccccc}
		\toprule
$\,$    & \textbf{E-PINN} & E-PINN(V) & B-PINN & DE \\
		\midrule
$\kappa$ & \textbf{1.00}  & 1.02
& 1.01 & 0.99 \\
$\sigma_\kappa$ & \textbf{0.01} & 0.01
& 0.04 & 0.09 \\
$\rho_e$ & \textbf{0.7} & 0.7 & 0.1 &
0.5 \\
$\rho_n$ & \textbf{0.8} & 0.6 & -0.2 &
0.5\\
ECP & \textbf{0.93} & 0.89 & 1.00 &
0.63 \\
$\Delta u_b$ & \textbf{0.002} & 0.006 & 0.01 & 0.001 \\
		\bottomrule
	\end{tabular}
	\label{tab:table2}
\end{table}
\section{Conclusion}
We have constructed a novel uncertainty-aware PINN-like model synergizing the fundamental principles of EDL and PINN. It provides another data-driven framework in scientific machine learning that is suited for solving inverse problems involving differential equations. E-PINN generates uncertainty estimates for \emph{both} the target variable and unknown PDE parameters
upon completion of model training. 
The novel information-theoretic regularizer we  proposed in eqn. \eqref{kl} appeared to enhance the model's effectiveness relative to the original version first proposed in \cite{amini}.  
For the 1D and 2D problems examined here, we found that E-PINNs superseded B-PINNs and other models in terms of sensitivity of uncertainties towards data noise and predictive errors. Its empirical coverage probability values were close to nominal ones indicating a good degree of uncertainty calibration. The 1D problem showed E-PINN to be effective in interpreting out-of-domain data, while the 2D problem illustrated the model's ability to adhere to boundary conditions.
An important future direction would be to validate E-PINN against a wider set of PDEs used in various scientific and engineering fields, such as those proposed in \emph{PINNacle} \cite{pinnacle}.


\bibliographystyle{acm}
\bibliography{Paper.bib}  

\begin{thebibliography}{10}

\bibitem{abdar}
{\sc Abdar, M., Pourpanah, F., Hussain, S., Rezazadegan, D., Liu, L., Ghavamzadeh, M., Fieguth, P., Cao, X., Khosravi, A., Acharya, U.~R., Makarenkov, V., and Nahavandi, S.}
\newblock A review of uncertainty quantification in deep learning: Techniques, applications and challenges.
\newblock {\em Information Fusion 76\/} (dec 2021), 243--297.

\bibitem{amini}
{\sc Amini, A., Schwarting, W., Soleimany, A., and Rus, D.}
\newblock Deep evidential regression.
\newblock In {\em Proceedings of the 34th International Conference on Neural Information Processing Systems\/} (Red Hook, NY, USA, 2020), NIPS '20, Curran Associates Inc.

\bibitem{pinnPy}
{\sc Bafghi, R.~A., and Raissi, M.}
\newblock Comparing {PINN}s across frameworks: {JAX}, tensorflow, and pytorch.
\newblock In {\em ICLR 2024 Workshop on AI4DifferentialEquations In Science\/} (2024).

\bibitem{baty2024}
{\sc Baty, H.}
\newblock A hands-on introduction to physics-informed neural networks for solving partial differential equations with benchmark tests taken from astrophysics and plasma physics, 2024.

\bibitem{Chen_2020}
{\sc Chen, Y., Lu, L., Karniadakis, G.~E., and Dal~Negro, L.}
\newblock Physics-informed neural networks for inverse problems in nano-optics and metamaterials.
\newblock {\em Optics Express 28}, 8 (Apr. 2020), 11618.

\bibitem{cobb}
{\sc Cobb, A.~D., and Jalaian, B.}
\newblock Scaling hamiltonian monte carlo inference for bayesian neural networks with symmetric splitting.
\newblock {\em Uncertainty in Artificial Intelligence\/} (2021).

\bibitem{thomas}
{\sc Cover, T.~M., and Thomas, J.~A.}
\newblock {\em Elements of Information Theory (Wiley Series in Telecommunications and Signal Processing)}.
\newblock Wiley-Interscience, USA, 2006.

\bibitem{cuomo2022}
{\sc Cuomo, S., di~Cola, V.~S., Giampaolo, F., Rozza, G., Raissi, M., and Piccialli, F.}
\newblock Scientific machine learning through physics-informed neural networks: Where we are and what's next, 2022.

\bibitem{Egele}
{\sc Egele, R., Maulik, R., Raghavan, K., Lusch, B., Guyon, I., and Balaprakash, P.}
\newblock Autodeuq: Automated deep ensemble with uncertainty quantification.
\newblock In {\em 2022 26th International Conference on Pattern Recognition (ICPR)\/} (Los Alamitos, CA, USA, aug 2022), IEEE Computer Society, pp.~1908--1914.

\bibitem{gal}
{\sc Gal, Y., and Ghahramani, Z.}
\newblock Dropout as a bayesian approximation: Representing model uncertainty in deep learning.
\newblock In {\em Proceedings of The 33rd International Conference on Machine Learning\/} (New York, New York, USA, 20--22 Jun 2016), M.~F. Balcan and K.~Q. Weinberger, Eds., vol.~48 of {\em Proceedings of Machine Learning Research}, PMLR, pp.~1050--1059.

\bibitem{DE}
{\sc Ganaie, M., Hu, M., Malik, A., Tanveer, M., and Suganthan, P.}
\newblock Ensemble deep learning: A review.
\newblock {\em Engineering Applications of Artificial Intelligence 115\/} (Oct. 2022), 105151.

\bibitem{ECP2}
{\sc Gneiting, T., Balabdaoui, F., and Raftery, A.~E.}
\newblock Probabilistic forecasts, calibration and sharpness.
\newblock {\em Journal of the Royal Statistical Society Series B: Statistical Methodology 69}, 2 (03 2007), 243--268.

\bibitem{ECP}
{\sc Gneiting, T., and Raftery, A.~E.}
\newblock Strictly proper scoring rules, prediction, and estimation.
\newblock {\em Journal of the American Statistical Association 102}, 477 (03 2007), 359--378.

\bibitem{hao2023}
{\sc Hao, Z., Liu, S., Zhang, Y., Ying, C., Feng, Y., Su, H., and Zhu, J.}
\newblock Physics-informed machine learning: A survey on problems, methods and applications, 2023.

\bibitem{pinnacle}
{\sc Hao, Z., Yao, J., Su, C., Su, H., Wang, Z., Lu, F., Xia, Z., Zhang, Y., Liu, S., Lu, L., and Zhu, J.}
\newblock Pinnacle: A comprehensive benchmark of physics-informed neural networks for solving pdes, 2023.

\bibitem{universal}
{\sc Hornik, K., Stinchcombe, M., and White, H.}
\newblock Multilayer feedforward networks are universal approximators.
\newblock {\em Neural Networks 2}, 5 (1989), 359--366.

\bibitem{Ameya}
{\sc Jagtap, A.~D., Mao, Z., Adams, N., and Karniadakis, G.~E.}
\newblock Physics-informed neural networks for inverse problems in supersonic flows.
\newblock {\em Journal of Computational Physics 466\/} (2022), 111402.

\bibitem{Jones}
{\sc Jones, C.~K., Wang, G., Yedavalli, V., and Sair, H.}
\newblock Direct quantification of epistemic and aleatoric uncertainty in 3d u-net segmentation.
\newblock {\em J Med Imaging (Bellingham) 9}, 3 (May 2022), 034002.

\bibitem{kim}
{\sc Kim, D., and Lee, J.}
\newblock A review of physics informed neural networks for multiscale analysis and inverse problems.
\newblock {\em Multiscale Science and Engineering 6\/} (02 2024).

\bibitem{Adam}
{\sc Kingma, D.~P., and Ba, J.}
\newblock Adam: A method for stochastic optimization, 2017.

\bibitem{Kullback}
{\sc Kullback, S., and Leibler, R.~A.}
\newblock {On Information and Sufficiency}.
\newblock {\em The Annals of Mathematical Statistics 22}, 1 (1951), 79 -- 86.

\bibitem{Lagaris_1998}
{\sc Lagaris, I., Likas, A., and Fotiadis, D.}
\newblock Artificial neural networks for solving ordinary and partial differential equations.
\newblock {\em IEEE Transactions on Neural Networks 9}, 5 (1998), 987–1000.

\bibitem{lak}
{\sc Lakshminarayanan, B., Pritzel, A., and Blundell, C.}
\newblock Simple and scalable predictive uncertainty estimation using deep ensembles.
\newblock In {\em Proceedings of the 31st International Conference on Neural Information Processing Systems\/} (Red Hook, NY, USA, 2017), NIPS'17, Curran Associates Inc., p.~6405–6416.

\bibitem{autograd}
{\sc Paszke, A., Gross, S., Chintala, S., Chanan, G., Yang, E., DeVito, Z., Lin, Z., Desmaison, A., Antiga, L., and Lerer, A.}
\newblock Automatic differentiation in pytorch.
\newblock In {\em NIPS 2017 Workshop on Autodiff\/} (2017).

\bibitem{pinn}
{\sc Raissi, M., Perdikaris, P., and Karniadakis, G.}
\newblock Physics-informed neural networks: A deep learning framework for solving forward and inverse problems involving nonlinear partial differential equations.
\newblock {\em Journal of Computational Physics 378\/} (2019), 686--707.

\bibitem{raissi2017}
{\sc Raissi, M., Perdikaris, P., and Karniadakis, G.~E.}
\newblock Physics informed deep learning (part i): Data-driven solutions of nonlinear partial differential equations, 2017.

\bibitem{sensoy}
{\sc Sensoy, M., Kaplan, L., and Kandemir, M.}
\newblock Evidential deep learning to quantify classification uncertainty, 2018.

\bibitem{Tan_2}
{\sc Tan, H.~S., Wang, K., and McBeth, R.}
\newblock Deep evidential learning for radiotherapy dose prediction.
\newblock {\em Computers in Biology and Medicine 182\/} (Nov. 2024), 109172.

\bibitem{bpinn}
{\sc Yang, L., Meng, X., and Karniadakis, G.~E.}
\newblock B-pinns: Bayesian physics-informed neural networks for forward and inverse pde problems with noisy data.
\newblock {\em Journal of Computational Physics 425\/} (Jan. 2021), 109913.

\bibitem{zhou2024}
{\sc Zhou, W., and Xu, Y.~F.}
\newblock Data-guided physics-informed neural networks for solving inverse problems in partial differential equations, 2024.

\bibitem{Zou}
{\sc Zou, K., Yuan, X., Shen, X., Wang, M., and Fu, H.}
\newblock Tbrats: Trusted brain tumor segmentation.
\newblock In {\em Medical Image Computing and Computer Assisted Intervention – MICCAI 2022: 25th International Conference, Singapore, September 18–22, 2022, Proceedings, Part VIII\/} (Berlin, Heidelberg, 2022), Springer-Verlag, p.~503–513.

\end{thebibliography}

\appendix 
\section{Appendix A: Supplementary derivations for Section \ref{Prelim}. }
In this appendix, we collect various
mathematical details omitted in Section \ref{Prelim} for additional clarity. 

To derive the residual function $\mathcal{G}$ in eqn. \eqref{simple_resi}, we note that
\begin{eqnarray}
\mathcal{G} &=& \int_{\mathcal{D}_\kappa} d\kappa \,\, \mathcal{P}\left( \kappa; \vec{\mu}_\kappa \right) \,\, \left[ 
\mathcal{L}^2_1 + \mathcal{L}^2_2 \kappa^2 + 
2\kappa \mathcal{L}_1 \mathcal{L}_2
\right] \cr
&=& \mathcal{L}^2_1 + \mathcal{L}^2_2 \overline{\kappa^2} + 
2\overline{\kappa} \mathcal{L}_1 \mathcal{L}_2 \cr
&\equiv& \mathcal{L}^2_1 + \mathcal{L}^2_2 (\text{Var}(\kappa) + \overline{\kappa}^2 ) + 
2\overline{\kappa} \mathcal{L}_1 \mathcal{L}_2.
\end{eqnarray}
When $\mathcal{G}$ is incorporated in Monte-Carlo Dropout for PINN, the uncertainty in $\kappa$ is not affected by the stochastic dropout units that are switched on during each forward pass, but has already been learned during model training. On the other hand, for Deep Ensemble, one averages over the collection of models to deduce both $\overline{\kappa}$ and Var($\kappa$). 

With regards to our regularizer loss term $\mathcal{L}_{kl}$, we had obtained it by taking 
the KL divergence between the normal-inverse-gamma (NIG) distribution parametrized by $\{\gamma, \alpha, \beta, \nu \}$
and another reference NIG distribution 
with $\{\gamma_r, \alpha_r, \beta_r, \nu_r \}$ that presumably characterizes a non-/weakly informative prior. When multiplied to a term such as $|u_i - \gamma|$ in \eqref{kl}, this means that for more significantly large errors, the loss term guides the distributions for such samples to approach a weakly-informative one yielding higher uncertainty. 
By definition, the KL divergence between two NIG distributions with density functions $(\tilde{p}_{r}, \tilde{p})$ is
\begin{equation}
D_{KL}\left( \tilde{p}_{r} \vert| \tilde{p}\right) = \iint \,d\overline{u} \,d\sigma^2_u \,\, \tilde{p}_{r} \log
\frac{\tilde{p}_{r}}{\tilde{p}}.
\end{equation}
The NIG distribution can be equivalently understood as the product 
$$
\overline{u} \sim \mathcal{N}
(\gamma, \frac{\sigma^2_u}{\nu} ), \,\,\, 
\sigma^2_u \sim \Gamma^{-1}(\alpha,\beta).
$$
We would like the reference prior to be a non- or weakly-informative prior. Taking the limit of $\nu_r \rightarrow \infty$ yields a uniform distribution over $\mathbb{R}$ which is unfortunately unnormalizable. To avoid this pathology, we take the reference NIG to be parametrized by values of $\gamma, \nu$ identical to those of the EDL model outputs' distributions, and instead set the reference inverse-gamma distribution to be of relatively higher entropy, taking $\alpha_r =1$ and letting $\beta_r$ be task-dependent. Since the variance $\sigma^2_u$
is not physically expected to be larger than the upper bound for $u$, a natural candidate for $\beta_r$ is the order-of-magnitude of $u_{max}$ --  the maximum value of the dependent variable as deduced from physical principles. For the differential equations examined in our work here, we take $\beta_r = 1$. Generally, it should be set to the same order-of-magnitude as $u_{max}$.  
This definition of $\tilde{p}_r$ leads to the KL divergence term being 
\begin{equation}
\label{div_kl}
\mathcal{D}_{KL}=
\left(
\log \frac{\beta^\alpha_r \, \Gamma(\alpha)} {\beta^\alpha} + \gamma_e (\alpha - 1)
+ \frac{\beta- \beta_r}{\beta_r}
\right).
\end{equation}
In this form, we note that there is no dependence on $\nu$. In \cite{amini}, the authors argued that the NIG distribution admits the interpretation of having its variance estimated from $\alpha$ `virtual observations' with sample mean $\gamma$ and sum of squared deviations $2\nu$, and one can thus interpret $2\nu + \alpha$ as a measure of evidence for the EDL distributions. Together with a linear term $|u_i -\gamma|$ that aims to increase the loss for predictions with larger errors, the authors of \cite{amini} then took 
$|u_i - \gamma|(2\nu + \alpha)$ to be the regularizer term. Since previous works \cite{amini, Tan_2} has shown it to be effective in a number of contexts and it carries a dependence on $\nu$, we multiply it to the KL term in
\eqref{div_kl} to obtain our final expression for the regularizer loss term
\begin{equation}
\mathcal{L}_{kl} = |u_i - \gamma|(2\nu + \alpha) D_{kl}.
\end{equation}

\section{Appendix B: Learning curves for E-PINN training}
\begin{figure}[H]
\small
	\begin{center}		\includegraphics[width=0.49\textwidth]{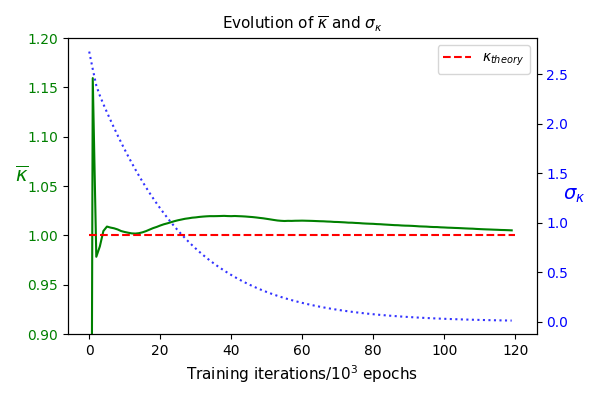}
	\end{center}
	\caption{\small Evolution of $\overline{\kappa}, \sigma_\kappa$ superposed on the same graph for the 2D case. }
	\label{fig:AppC_para}
\end{figure}

Model training for E-PINN involves
choices of loss function weight coefficients $\lambda_d, \lambda_r, \lambda_{kl}$. Apart from the individual loss terms, they should yield learning curves which show stable convergent behavior for $\overline{\kappa}, \sigma_\kappa$. Fig. \ref{fig:AppC_para}
depicts the evolutions of $\overline{\kappa}, \sigma_\kappa$ for the 2D problem. 
While the latter decreased steadily, the former experienced a transient oscillatory phase before becoming stable near the theoretical $\kappa$ value. This phase mirrored the dynamical balance between the PDE residual loss $\mathcal{G}$ and the EDL-type loss terms 
$\lambda_d \mathcal{L}_{edl} + \lambda_{kl} \mathcal{L}_{kl}$.

\begin{figure}[H]
	\begin{center}		\includegraphics[width=0.49\textwidth]{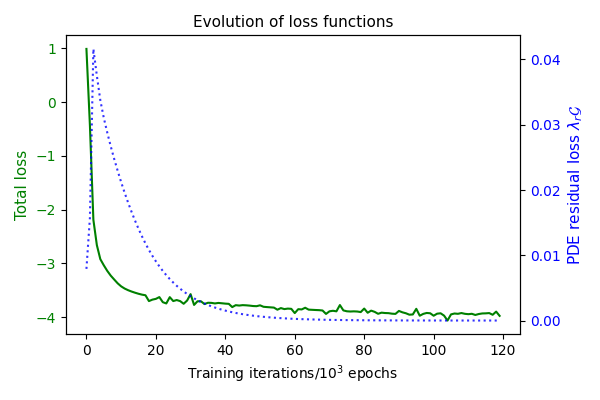}
	\end{center}
	\caption{\small Evolution of the total and PDE loss functions for the 2D case. }
	\label{fig:AppC_loss}
\end{figure}

In Fig. \ref{fig:AppC_loss}, we observe that the early sharp rise in the PDE residual loss parallels the transient oscillatory learning curve of $\overline{\kappa}$ present in Fig. \ref{fig:AppC_para}. The EDL data loss term $\lambda_d \mathcal{L}_{edl} + \lambda_{kl} \mathcal{L}_{kl}$ evolved similarly to the total loss function shown in Fig. \ref{fig:AppC_para}.
Similar trends for these training curves and evolutions of $\overline{\kappa}, \sigma_\kappa$ were observed for the 1D case too. 
Setting $\lambda_d = 1$, we 
took $\lambda_{kl} = 0.01, \lambda_r = 0.1$ for the 1D problem and 
$\lambda_{kl} = 0.05, \lambda_r = 0.5$ for the 2D problem. Like the usual PINN framework, E-PINN training is fundamentally semi-supervised in nature, blending a data-based component with a PDE-residual term, with no labels in the training dataset for $\kappa, \sigma_\kappa$ nor the uncertainty estimates. Yet we can use the degree of convergent behaviors of the each loss function component and evolution of $\kappa, \sigma_\kappa$ to guide us in excluding unsuitable values of $\lambda_d,\lambda_{kl}, \lambda_r$. In more realistic application settings, the uncertainties in the dependent variable $u$ and unknown parameter $\kappa$ may be linked to physical principles (such as the precision of sensors measuring $\kappa$ or $u$) which would give an order-of-magnitude guide for $\sigma_\kappa, \sigma_p$. This in turn can be useful for picking suitable weight coefficients for each loss term.

\section{Appendix C: Additional details on the results for other uncertainty-aware PINN models}
\label{sec:AppC}
In this Appendix, we provide some additional details on the simulation and training results for 
\renewcommand{\labelenumi}{[\Alph{enumi}]}
\begin{enumerate}
\item B-PINN with Hamiltonian-Monte-Carlo for posterior sampling \cite{bpinn}.
\item E-PINN(V): E-PINN equipped not with \eqref{kl} but with the original regularizer of \cite{amini} : $\mathcal{L}_{kl} = |u_i - \gamma| (2\nu + \alpha)$,
\item Deep Ensemble (DE) method using a set of 20 PINNs with distinct initial weights.
\end{enumerate}

\begin{figure}[H]
	\begin{center}		\includegraphics[width=0.48\textwidth]{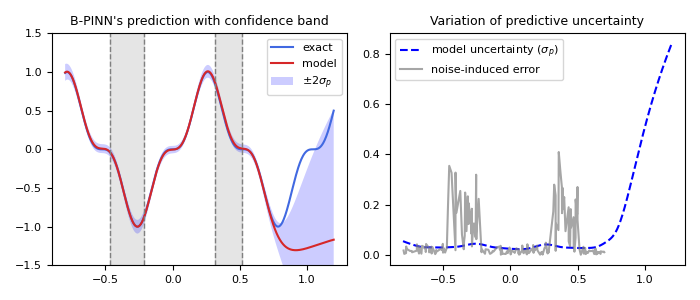}
	\end{center}
	\caption{\small B-PINN's predictions and uncertainty distributions (1D problem).}
	\label{fig:AppC_0}
\end{figure}

\textbf{[A]} B-PINN models exhibited weaker correlations with noise and errors compared to our E-PINN models. 
In Fig. \ref{fig:AppC_0}, one can see that the confidence band was not particularly enlarged within the high-noise regions, in contrast to our E-PINN model (Fig. \ref{fig:fig3}). 
It yielded weak correlations between $\sigma_p$ and model errors $(\rho_e \sim 0.5)$ or noise-induced errors $(\rho_n \sim 0.2)$.
The model's empirical coverage probability was 0.87. Predicted value for $\kappa$ with its uncertainty was $\kappa = 0.69 \pm 0.03$, while its mean predictive uncertainty was $\sigma_p \sim 0.03$. 

For the 2D problem, 
B-PINN model yielded
uncertainty estimates that tended to be inflated relative to magnitudes of model errors (ECP was thus unity) as depicted in Fig. \ref{fig:AppC_0}, with vague correlations between $\sigma_p$ and model errors $(\rho_e \sim 0.1)$ or noise-induced errors $(\rho_n \sim -0.2)$. The mean uncertainty $\overline{\sigma_p} \sim 0.02$. For the PDE parameter, the model inferred $\kappa \sim 1.01 \pm 0.04$.

\begin{figure}[H]
	\begin{center}		\includegraphics[width=0.49\textwidth]{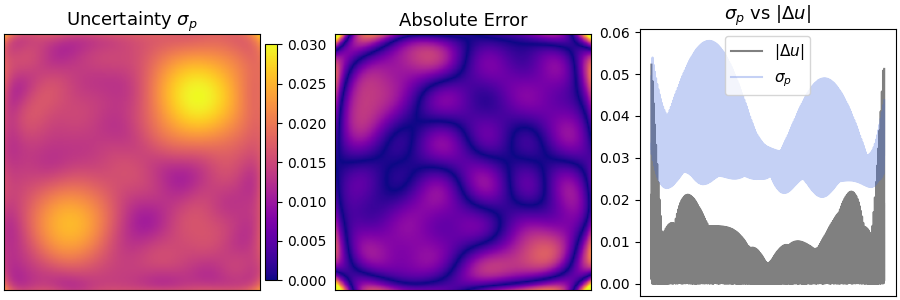}
	\end{center}
	\caption{\small B-PINN's predictions and uncertainty distribution heatmaps (2D problem). }
	\label{fig:AppC_1}
\end{figure}

\begin{figure}[H]
	\begin{center}		\includegraphics[width=0.49\textwidth]{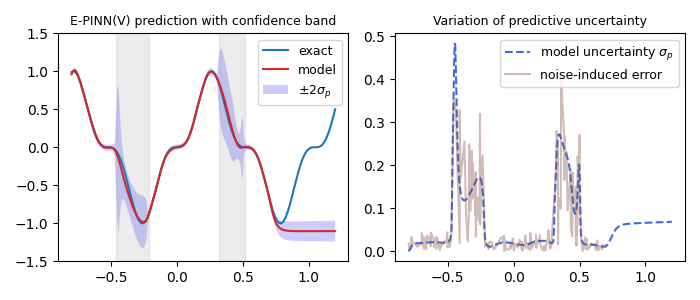}
	\end{center}
	\caption{\small Predictions and uncertainty distributions for E-PINN(V) (1D problem).}
	\label{fig:AppC_2}
\end{figure}

\textbf{[B]} For E-PINN(V), we found that the main disadvantage relative to our proposed 
$\mathcal{L}_{kl}$ was a weaker response to out-of-distribution data. In Fig. \ref{fig:AppC_2}, one can see that although the confidence band was enlarged within the high-noise regions, in contrast to our E-PINN model (Fig. \ref{fig:fig3}), the predictive uncertainty $\sigma_p$ was relatively much lower and not encompassing the deviations from the exact solution. The Spearman's coefficients $\rho_s$ were $\{ 0.6, 0.7 \} $ for uncertainty correlations with error and noise respectively. Its empirical coverage probability was 0.79. Predicted value for $\kappa$ with its uncertainty was $\kappa = 0.74 \pm 0.03$, while its mean predictive uncertainty was $\sigma_p \sim 0.02$. 
For the 2D problem, E-PINN(V) was weaker than our model in the aspect of the inferred parameter's accuracy, ECP value and mean error along boundary, yet was comparable in other aspects tabulated in Table \ref{tab:table2}. 

\begin{figure}[H]
	\begin{center}		\includegraphics[width=0.49\textwidth]{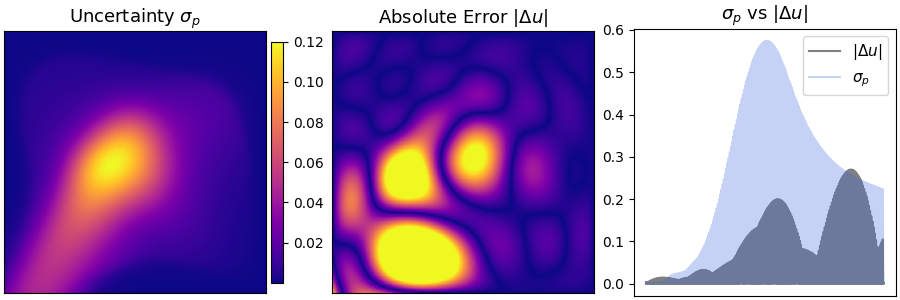}
	\end{center}
	\caption{\small E-PINN(V)'s predictions and uncertainty distribution heatmaps (2D problem).}
	\label{fig:AppC_3}
\end{figure}

\textbf{[C]} We implemented DE
using a set of 20 PINNs with $\kappa$ as the only trainable parameter in the PDE residual loss), with distinct initial random weights set in Pytorch via Kaiming initialization. Training was completed with $10^5$ epochs with Adam optimizer under a constant learning rate of $10^{-3}$. For the 1D problem
DE generated an uncertainty spectrum that did not appear to be sensitive to noise at all, though its confidence band for out-of-domain data was better calibrated compared to B-PINN and E-PINN(V). 
Predicted value for $\kappa$ with its uncertainty was $\kappa = 0.6976 \pm 10^{-4}$. 

\begin{figure}[H]
	\begin{center}		\includegraphics[width=0.49\textwidth]{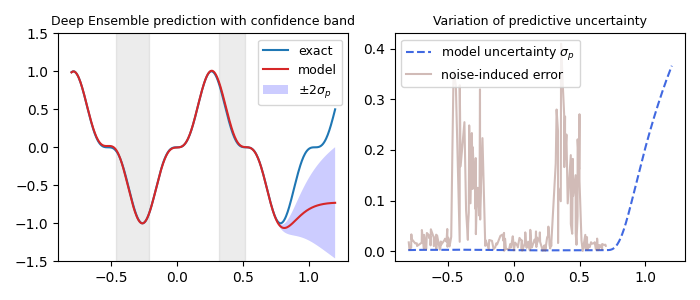}
	\end{center}
	\caption{\small Predictions and uncertainty distributions for the Deep Ensemble method (1D problem).}
	\label{fig:AppC_4}
\end{figure}

For the 2D problem, DE inferred the PDE parameter $\kappa \sim 0.99 \pm 0.09$, and relative to its performance in the 1D case, yielded a better ECP = 0.63. The mean uncertainty was $\overline{\sigma_p} \sim 0.003$, with the mean error $\sim 0.008$ error restricted to boundary being $\sim 0.001$.
Relative to B-PINN, it notably had better uncertainty-error correlations, with $\rho_e, \rho_n$ both being $\sim 0.5$. 
From both 1D and 2D problems, the main weakness of this method seemed to be that the uncertainty values were too low giving rise to a rather problematic ECP. This implied that DE did not yield well-calibrated uncertainty estimates, as was also apparent in both uncertainty distribution plots.

\begin{figure}[H]
	\begin{center}		\includegraphics[width=0.49\textwidth]{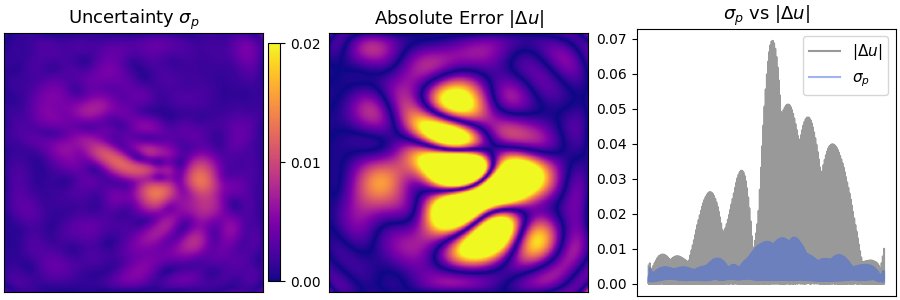}
	\end{center}
	\caption{\small Predictions and uncertainty heatmaps for the Deep Ensemble method (2D problem).}
	\label{fig:AppC_5}
\end{figure}

\end{multicols}
\end{document}